\documentclass[a4paper, 10pt, conference]{ieeeconf}      
\usepackage{graphicx}
\usepackage{amsmath,amsfonts,amssymb}
\usepackage{bm}
\usepackage{comment}
\usepackage{subfigure}
\usepackage{siunitx}
\newcommand{\mvect}[1]{\mathbf{#1}}

\newenvironment{rcases}
{\left.\begin{aligned}}
	{\end{aligned}\right\rbrace}

\IEEEoverridecommandlockouts                              
\title{\LARGE \bf
Automated design of pneumatic soft grippers through design-dependent multi-material topology optimization 
}

\author{Josh Pinskier$^{1}$, Prabhat Kumar$^{2}$, Matthijs Langelaar$^{3}$, and David Howard$^{1}$
\thanks{$^{1}$Robotics and Autonomous Systems Group, CSIRO Data61, Brisbane, Australia
        {\tt\small josh.pinskier@csiro.au, david.howard@csiro.au}}%
\thanks{$^{2}$Department of Mechanical and Aerospace Engineering, Indian Institute of Technology Hyderabad, 502285 Telangana, India
        {\tt\small pkumar@mae.iith.ac.in}}%
\thanks{$^{3}$Department of Precision and Microsystems Engineering, TU Delft, Delft, Netherlands
        {\tt\small m.langelaar@tudelft.nl}}%
\thanks{The authors thank Prof. Krister Svanberg for providing MATLAB codes of the MMA optimizer.}
}

\begin{document}
\maketitle
\thispagestyle{empty}
\pagestyle{empty}

\begin{abstract}
Soft robotic grasping has rapidly spread through the academic robotics community in recent years and pushed into industrial applications. At the same time, multimaterial 3D printing has become widely available, enabling the monolithic manufacture of devices containing rigid and elastic sections. We propose a novel design technique that leverages both technologies and can automatically design bespoke soft robotic grippers for fruit-picking and similar applications. We demonstrate the novel topology optimisation formulation that generates multi-material soft grippers, can solve internal and external pressure boundaries, and investigate methods to produce air-tight designs. Compared to existing methods, it vastly expands the searchable design space while increasing simulation accuracy.
\end{abstract}
  
\section{Introduction}
Soft robotic grasping has emerged as a safe and effective means for grasping fragile, flexible and fluctuating objects. Their inherent deformability enables them to conform to fit the objects' shape and distribute gripping force, hence gently grasping even soft objects. 

These soft grippers are often inspired by human hands, which are seen as the gold standard in soft and dexterous grasping. However, there is an increasing trend towards non-anthropomorphic designs, which enable diverse grasping strategies and require controllable fewer degrees of freedom (DOFs) \cite{Hao2021}. Several mechanisms have been investigated for their actuation including pneumatic \cite{Mosadegh2014}, tendon-driven \cite{Laschi2012}, and granular (vacuum) jamming \cite{Brown2010,Howard2021}.

\begin{figure}[h!]
\centering
\includegraphics[width=0.9\columnwidth]{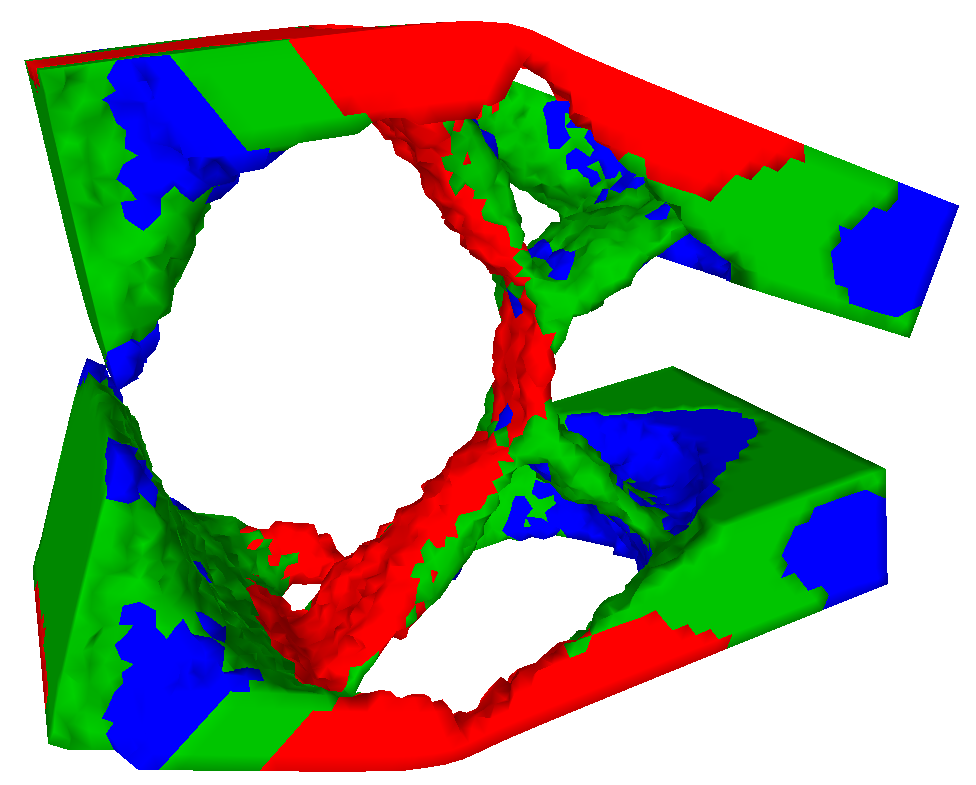}
\caption{3-Material optimised soft-gripper under 50kPa pressure (5x deformation scale). Pressure is applied to the two faces on the left, causing the jaws to close on the right. Material stiffnesses are: Red - 100MPa, Green - 10MPa, Blue - 1MPa}
\label{fig:Hero_SoftGripper}
\end{figure}

Despite the diversity of grasping and actuation paradigms available in the literature, most existing grippers are hand-designed. They draw on human experience and biomimicry to navigate the complexity of designing deformable devices to generate high-quality designs \cite{Pinskier2022}. The resulting generic designs emphasise universal approaches rather than bespoke designs \cite{Brown2010}.
However, real-world applications frequently require designs which are tailored to the specifics of their task. Clearly, a fruit-picking robot requires a different end effector to an assembly line robot or a human assistance robot, and an apple-picking end-effector has different requirements to a strawberry picker.

Despite this obvious need to produce bespoke soft end-effectors, existing automated design tools are limited and underexplored. Methods including simulated and in-materio evolution have recently proven successful in designing granular jamming grippers \cite{Howard2022a, Fitzgerald2021}. More generally, an impressive and diverse array of soft robotics have been evolved in simulation \cite{Hiller2009, Auerbach2011}. However, because of the large number of evaluations required these require very cheap simulators, which are unable to capture multiphysics interactions and have a large simulation to reality gap \cite{Kriegman2020a}.

In contrast, Topology optimization (TO) is a general purpose design tool, suitable to numerous actuation techniques and physical domains \cite{sigmund2013topology}. It distributes material inside a meshed (or pixelized/voxelized) space to identify the topology with the best performance, and has designed both pneumatic and tendon-driven soft grippers. \cite{Chen2018a,Pinskier2022,kumar2020topology3Dpressure}. 
However, the methods presented in these works require significant assumptions about the design domain and actuation, limiting both the accuracy of the simulation and the range of realisable designs. 

\subsection{Topology Optimisation of Soft Grippers}
The current state-of-the-art in topology optimised soft grippers broadly falls into two categories. Either externally actuated grippers which use an exogenous displacement to drive their grasping behaviour (an externally routed cable or moving surface) \cite{Liu2018,Chen2018a}; or pressure-actuated soft fingers without design dependency \cite{Zhang2019,Liu2022}. In both cases, the actuation source is prespecified and does not form part of the optimisation problem. Whilst convenient, these assumptions do not reflect best-practice design methods which use complex pneumatic chambers and internal cable routing. 
To capture these features, the loading point (magnitude, direction and location) should be free to move with each iteration of the topology optimisation solver. This design-dependency problem increases the solver complexity and requires auxiliary physics equations to solve and additional constraints to enforce physical limits. A small number of topology optimised soft grippers have investigated design-dependent pressure optimisation, but their coarse physics approximations result in unrealisable designs, with disconnected pressurised regions \cite{Chen2019a,Caasenbrood2020}.

The above methods have been demonstrated only in single-material optimisation. However, improvements in 3D printing technology, enable the monolithic manufacture of arbitrarily complex multi-material soft robots. With two or more materials it becomes possible to strike a trade-off between the flexibility and strength of the material, and increase the overall strength of the device without compromising on its workspace. For a detailed review of soft robotic topology optimisation see \cite{Pinskier2022, Chen2020}. To the best of the authors' knowledge, there is currently no method for creating multi-material pneumatically activated soft robots using topology optimization. 

\subsection{Pressure-Loaded Topology Optimsation}
Pressure-loaded topology optimsation is a problem that extends beyond soft robotics. It has applications in the design of pneumatically and hydraulically loaded structures like pressure vessels, dams, pumps and ships.
In these problems, the fluid-solid boundary and hence the loading must move during the optimisation. In density-based topology optimisation, mesh elements are allowed to occupy a continuum between solid and void \cite{sigmund2013topology}. Hence, the problem is commonly approached either by attempting to explicitly identify a fluid-solid boundary, or using a mixed fluid-solid formulation \cite{Sigmund2007a,kumar2020topology}. In contrast, the current state of the art method treats the continuous density material as a porous media, and uses the Darcy method to estimate fluid penetration as a function of density \cite{kumar2020topology, kumar2020topology3Dpressure}. This allows the boundary to be located implicitly without the need to explicitly seperately the fluid-filled and void regions. However, generating airtight in pressure-actuated compliant mechanisms remains challenging as the contiguous, closed surfaces required to hold pressure also reduce compliance.
Hence, the problem and cost function design are critical to prevent leaky designs. 
The issue can be resolved using a material filtering scheme, which forces a solid layer between the high and low pressure regions \cite{Desouza2020}, such a scheme is heavily dependent on the optimiser's initial conditions and prevents the formation of beneficial internal cavities. 

\subsection{Contributions}
In this work, we present a novel method to design 3D multi-material pressure-actuated soft grippers using topology optimization. The method builds on our previous work into pressure-loaded topology optimization using Darcy's law \cite{kumar2020topology,kumar2020topology3Dpressure} and the extended solid-isotropic material with penalisation (SIMP) material model for the multi-material modeling~\cite{sigmund1997design}. An example of a soft gripper designed using this method is shown in Figure \ref{fig:Hero_SoftGripper}, it uses three materials with stiffnesses of $1~MPa$, $10~MPa$ and $100~MPa$. Using the multimaterial Darcy formulation, the solver converges to a soft gripper which clamps together using several compliant hinges. 
The main contributions of this work are:
\begin{enumerate}
    \item The first presentation of a multi-material topology optimsation formulation for pneumatic soft robots with design-dependent loading conditions.
    \item The development and investigation two new formulations to generate sealed pneumatic actuators, based on pressure regions and an energy penalty, respectively.
    \item The design of several new multimaterial pressure-actuated soft grippers.
\end{enumerate}
We focus on the application of this methology to soft robotic grasping, but it is generalisable to other pneumatic compliant mechanism and soft robots.

\section{Topology Optimisation Formulation}
\label{sec:topopt_formulation}
In this work, we use the density based SIMP method for topology optimisation. The goal of topology optimisation is to find a discrete material layout where each region contains a unique material or is left void. To simplify the problem, SIMP allows the design variable $\rho$ to occupy a continuum from 0 to 1, and a penalty $p=3$ applied to drive the results towards a binary solution. 
For a single material problem this is done using the SIMP interpolation law:
\begin{equation}~\label{Eq:twoMaterial}
	E_i = (1 -\bar{\rho}^p)E_\mathrm{min} + \bar{\rho}^pE_1
\end{equation}
where $E_\text{min}$ is a small, non-zero constant used to prevent singularities in material voids and $E_1$ is the elastic modulus of the material used.

\subsection{Multimaterial Modeling}\label{Sec:MultimaterialModeling}
We apply the extended SIMP interpolation technique to model multiple materials for the gripper mechanisms \cite{sigmund1997design}. In this formulation, one design variable is assigned to each material. For example, in the two-material case, the scheme with the modified SIMP formulation~ can be written as:
\begin{equation}~\label{Eq:twoMaterial}
	E_i = (1 -\bar{\rho}_{i1}^p)E_\mathrm{min} + \bar{\rho}_{i1}^p ((1-\bar{\rho}_{i2}^p)E_1 + \bar{\rho}_{i2}^p E_2)
\end{equation}
where $E_1$ and $E_2$ are  moduli of material~1 and material~2, respectively. $\bar{\rho_i}$ denotes the physical variable corresponding to design variable  $\rho_i$. $\left\{\bar{\rho}_{i1} =1,\,\bar{\rho}_{i2} =1\right\}$ gives the second material, whereas $\left\{\bar{\rho}_{i1} =1,\,\bar{\rho}_{i2} =0\right\}$ provides the first material. Thus, $\bar{\rho}_{i1}$ is called the topology variable. It decides the topology of the evolving design, whereas  $\bar{\rho}_{i2}$ decides the candidate material. 
Similarly the three-material case can be described by:
\begin{equation}~\label{Eq:threeMaterial}
	\begin{split}
		E_i = (1 -\bar{\rho}_{i1}^p)E_\mathrm{min} + \bar{\rho}_{i1}^p [(( 1-\bar{\rho}_{i2}^p )E_1 + 
        \\ \bar{\rho}_{i2}^p ((1-\bar{\rho}_{i3}^p)E_2 + \bar{\rho}_{i3}^p E_3))] 
	\end{split}
\end{equation}
where $E_3$ is the modulus of material~3. Using three materials, 
$\left\{\bar{\rho}_{i1}=1,\,\bar{\rho}_{i2}=0 ,\,\bar{\rho}_{i3}=0\right\}$, $\left\{\bar{\rho}_{i1}=1,\, \bar{\rho}_{i2}=1 ,\,\bar{\rho}_{i3}=0\right\}$ and $\left\{\bar{\rho}_{i1}=1,\, \bar{\rho}_{i2}=1 ,\,\bar{\rho}_{i3}=1\right\}$ give material~1, material~2 and material~3. 
To remove non-physical checkerboard patterns and intermediate (i.e non-binary) densities from the final design, we use a spacial density filter with hyperbolic projection as in \cite{sigmund2013topology,bruns2001topology}. This takes a weighted averages of the elemental density with its neighbours, then uses a hyperbolic projection to drive towards 0/1.


\subsection{Pressure load modeling}\label{Sec:PressLoadModel}
The method developed here for pneumatic soft robotic optimisations builds on our previous work into the Darcy method, a detailed description of which can be found in \cite{kumar2020topology,kumar2020topology3Dpressure}. It conceptualises the continuous design variable $\bar{\rho}$ as a porous medium, and uses Darcy's law to calculate pressure losses. In it, the flux $\bm{q}$ (volumetric fliud flow rate across a unit area) is defined by the flow coefficient $K(\bar{{\rho}}_{i1})$ and the pressure difference $\nabla p$ as:
\begin{equation}\label{Eq:Darcyflux}
	\bm{q} = -\frac{\kappa}{\mu}\nabla p = -K(\bar{{\rho}}_{i1}) \nabla p
\end{equation}
As the topology of the multimaterial structure (whether there is a material or void) is determined by $\bar{{\rho}}_{i1}$, the flux solely depends on $\bar{{\rho}}_{i1}$, regardless of the number of materials. Hence, the flow coefficient of element~$i$ is calculated as
\begin{equation}\label{Eq:Flowcoefficient}
	K(\bar{\rho}_i) = K_v\left(1-(1-\frac{K_s}{K_v}) \mathcal{H}(\bar{\rho}_{i1},\,\beta_\kappa,\,\eta_\kappa)\right)
\end{equation}
where
\begin{equation}
\mathcal{H}(\bar{{\rho}}_{i1},\,\beta_\kappa,\,\eta_\kappa) = \frac{\tanh{\left(\beta_\kappa\eta_\kappa\right)}+\tanh{\left(\beta_\kappa(\bar{\rho}_{i1} - \eta_\kappa)\right)}}{\tanh{\left(\beta_\kappa \eta_\kappa\right)}+\tanh{\left(\beta_\kappa(1 - \eta_\kappa)\right)}}
\end{equation}
$K_s$ and $K_v$ are flow coefficients of solid and void phases, respectively, and $\eta_\kappa$ and $\beta_\kappa$ shape the distribution of $K(\bar{\rho}_i)$.

A drainage term, $Q_\text{drain}$, is added. It helps achieve the natural pressure field variation by draining pressure from internal cavities:
\begin{equation}
{Q}_\text{drain} = -D_s \mathcal{H}(\bar{{\rho}_{i1}},\,\beta_d,\,\eta_d)(\bar{\rho}_e) (p - p_{\text{atm}})
\end{equation}
where $Ds$ is drainage coefficient and $p_{\text{atm}}$ is the atmospheric pressure. 
The net flow of the system is given by the equilibrium equation:
\begin{equation}\label{Eq:stateequation}
	\nabla\cdot\bm{q} -Q_\text{drain} =  0
\end{equation}
Which is solved using the finite element method to find the equilibrium pressure distribution and transform the pressure distribution $\mvect{p}$, to a global force $\mvect{F}$ to solve the mechanical equilibruim equation:
\begin{equation}\label{Eq:nodalforce}
	\mathbf{Ku}=\mvect{F} = -\mvect{T}\mvect{p}
\end{equation}
where $\mathbf{u}$ and $\mathbf{K}$ are the global displacement vector and stiffness matrix, and $\mvect{T}$ tranforms elemental pressures to nodal forces. A linear system is used to facilitate a tractible and efficient solution. However, the resulting solution is accurate only for small deformations.
By using two physical equation to solve for the equilibrium pressure and displacement, the formulation determines the pressure boundary at each iteration. 

\subsection{Problem formulation}\label{Sec:problem_formulation}
The final optimisation problem is formulated using:

 \begin{equation}\label{Eq:Optimizationequation}
\begin{rcases}
& \underset{\bm{\rho}}{\text{min}}
& &-s\frac{u_\text{out}}{(SE)^{1/n}}\\
& \text{such that:}  &&\,\, \mathbf{Ap} = \mathbf{0 }\\
&  &&\,\,\mathbf{Ku = F} = -\mathbf{T p}\\
&  && \,\,\displaystyle \sum_{i=1}^{\texttt{nel}}v_i\bar{\rho}_{i1}\le \left(v_{f_1} + v_{f_2} + v_{f_3}\right) \sum_{i=1}^{\texttt{nel}}v_i\\
&  && \displaystyle \sum_{i=1}^{\texttt{nel}}v_i\bar{\rho}_{i2}\le v_{f_2}\sum_{i=1}^{\texttt{nel}}v_i\\
&  && \displaystyle \sum_{i=1}^{\texttt{nel}}v_i\bar{\rho}_{i3}\le v_{f_3}\sum_{i=1}^{\texttt{nel}}v_i\\
&  && \mathbf{0}\le\bm{\bar{\rho}}\le \mathbf{1}\\
\end{rcases},
\end{equation}

where $u_\text{out}$ and $SE$ indicate output displacement and strain energy, respectively. $s$ is the consistent scaling parameter. $\mathbf{A}$ is the global flow matrix, which is found by assembling \eqref{Eq:stateequation}. We use three linear volume constraints using the definitions $\bar{\rho}_{i1}$, $\bar{\rho}_{i2}$ and $\bar{\rho}_{i3}$ described above. The first constraint controls the total amount of the solid state, whereas the second and third give the material amount of phase~2 and phase~3. $v_{f_1}$, $v_{f_2}$ and $v_{f_3}$ denote the volume fraction for material~1, material~2 and material~3, respectively. 

The cost function is selected to balance the dual requirements of maximising the deformation of the gripper, and maintaining a design which is stiff enough to grasp and hold objects. Here $n=8$ was selected after some initial studies to place a soft penalty on the design's stiffness. In this work, $v_{f_1}=0.3$, $v_{f_2}=0.2$, and $v_{f_3}=0.2$ unless otherwise stated, hence the total material permitted is 70\% of the volume of the design domain. Whilst it is desirable to minimise material usage, permitting more material is desirable for proof of concept. In each optimisation, the design variables are initialised with a constant density $\bar{\rho}_{in}=v_{f_n}$ for material $n$. Finally, the input pressure is $\SI{50}{\kilo\pascal}$ and the materials are given stiffnesses $E_1=\SI{1}{MPa}$, $E_2=\SI{10}{MPa}$, and $E_3=\SI{100}{MPa}$.


\section{Soft Grippers Design}
\label{sec:soft_gripper}
\begin{figure}[h!]
\centering
\subfigure[]{
\label{fig:gripper_designspace}
\includegraphics[width=0.45\columnwidth]{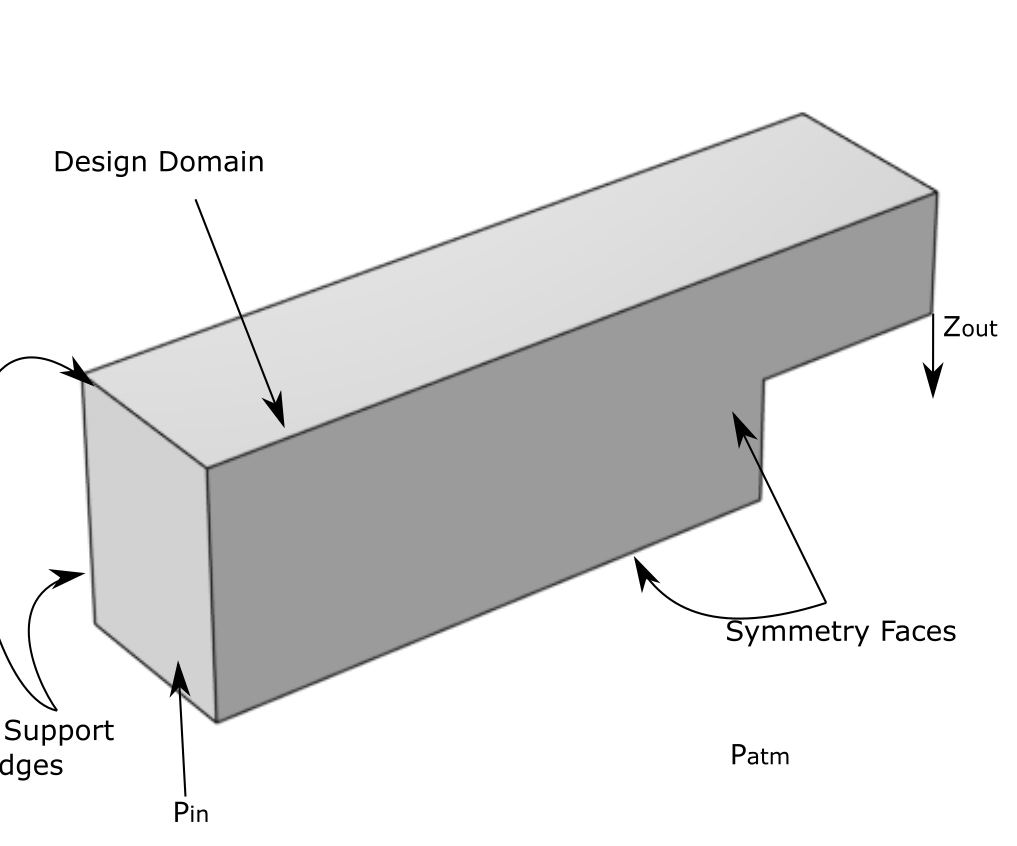}}
\subfigure[]{
\label{fig:gripper_undeformed}
\includegraphics[width=0.4\columnwidth]{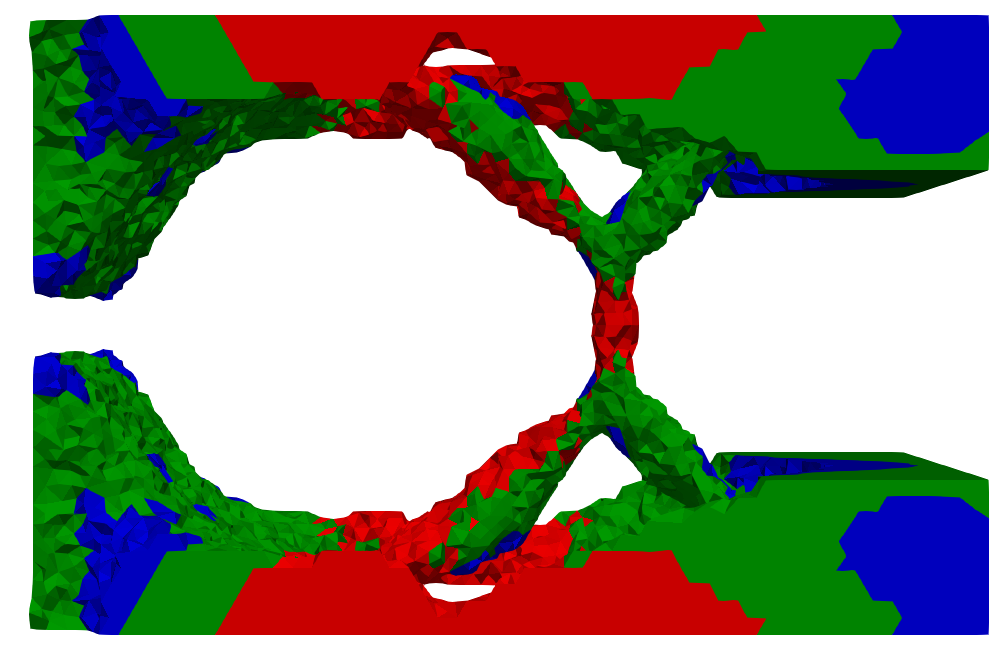}}
\subfigure[]{
\label{fig:gripper_undeformed2}
\includegraphics[width=0.4\columnwidth]{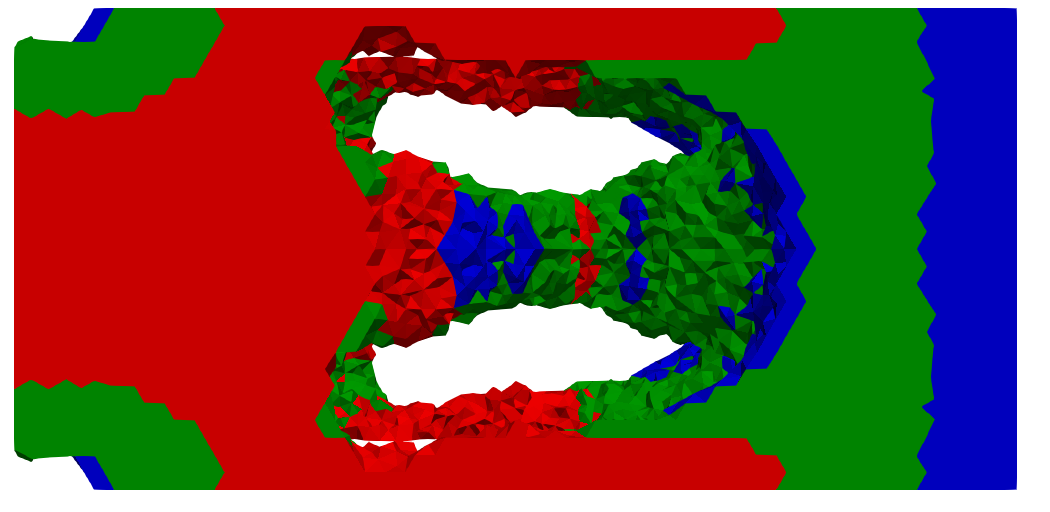}}
\subfigure[]{
\label{fig:gripper_deformed}
\includegraphics[width=0.4\columnwidth]{Figures/gripper00.png}}
\caption{3-Material optimised soft-gripper with stiffnesses: Red - 100MPa, Green - 10MPa, Blue - 1MPa,(a) Design domain  (b) Undeformed Side-view (c) Undeformed Top-View}
\label{fig:Soft-Gripper}
\end{figure}

To demonstrate the method and motivate the need for airtightness, this section investigates the design of pressure-actuated grippers using the multimaterial Darcy formulation.
The design domain of the grippers is presented in Figure \ref{fig:gripper_designspace}. Pressure is applied from the left face, with the output direction shown on the right. To simplify the domain and reduce computation time, 2 planes of symmetry are used, reducing workspace size.

The resulting design is illustrated in Figures \ref{fig:gripper_undeformed} and \ref{fig:gripper_undeformed2}, showing the design domain and undeformed configuration. The deformed configuration is show in Figure \ref{fig:gripper_deformed}.
In it, a solid face is formed on the left side, which absorbs the pressure. 
The internal strains are then transferred to the output face via a series of compliant hinges, one in the centre of the gripper, and four on the outer edges. Thin sections of the stiffest material $E_3$ are used in each hing, and joined by the softer materials $E_1$ and $E_2$.
Although quite elegant, the design illustrates two issues with existing pressure optimisation methods. The first is that the optimiser frequently falls into a local minimum in which the pressurised fluid is not allowed to penetrate deeply into the structure, preventing the formation of more complex, higher performing designs. The second is that without careful consideration of the design domain, the optimiser generates holes in the final design which spuriously increases performance by reducing stiffness in undesired locations.
In this case, resealing the device is fairly trivial, but in more complex designs, doing so adversely affects performance. Hence, design methods are needed which drive closed designs.

\section{Airtight Design}
\label{sec:Airtight_Problem}
To generate closed designs, we investigate and compare three methods, and apply them to soft finger design.
In soft fingers, the pressure load is often applied via a central channel in the design domain. This forces pressure deeper into the design and enhances performance but also increases its susceptibility to hole generation. Viewed from the perspective of the optimsation problem, sealed chambers reduce compliance and restrict deformation.
We propose two new methods for generating sealed designs:
\begin{enumerate}
    \item A heuristic approach, which adds material to the final design along the median pressure contour. 
    \item A penalty approach, which adds an energy term to the cost function and drives the optimisation to reduce pressure loss.
\end{enumerate}

The first approach leverages the advantages of the Darcy method, which calculates the internal pressure distribution between the inlet and outlet points. Where a face is unsealed, a smooth pressure gradient will flow from the inlet to the outlet. However, closed regions have a sharper pressure boundary. Hence by adding material along the line $0.5(P_{in}-P_{atm})$ we close open regions without significantly impacting regions which already have material.

The second approach is more rigorous, but remains susceptible to local minima. Using the equilibrium flow from the Darcy equation, we are able to calculate the energy transferred from inlet to outlet. In a closed system, there would be no flow, hence no energy transferred. However, using the Darcy method, a small flow will always arise. We use this energy value as a penalty term in the cost function, such that we seek to minimise:
\begin{equation}
\underset{\bm{\rho}}{\text{min}}
~~~-s\frac{u_\text{out}}{E_t(SE)^{1/n}}\\
\end{equation}
where $E_t$ is the total energy loss calculated at the boundaries and $s$ is a constant.  

\section{Airtight Soft Fingers}
\label{sec:Soft_Fingers}
The design domain of the soft fingers is presented in Figure \ref{fig:Soft-Finger-Design-space}. It is fixed around the edges on the left side and pressure enters via a central cavity, a single symmetry face is used to reduce the problem size. The aim is to maximise the bending on the right side.
\subsection{Heuristic Skin}
\begin{figure}[h!]
\centering
\subfigure[]{
\label{fig:Soft-Finger-Design-space}
\includegraphics[width=0.45\columnwidth]{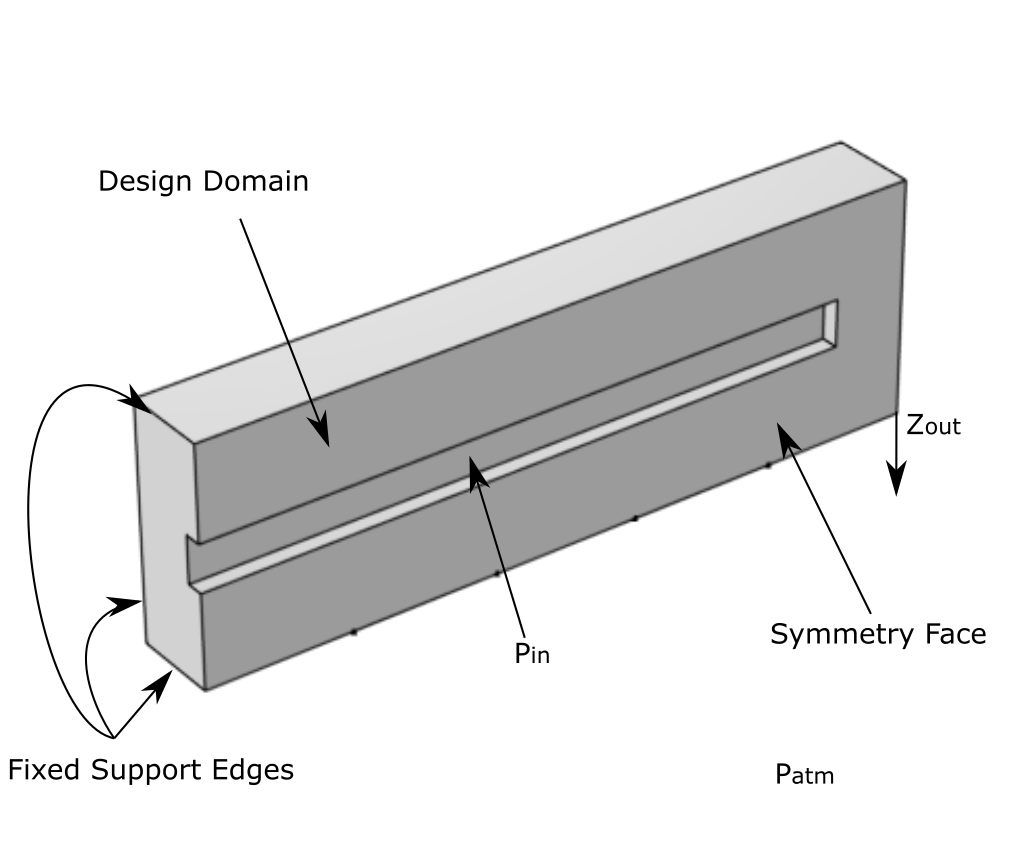}}
\subfigure[]{
\includegraphics[width=0.45\columnwidth]{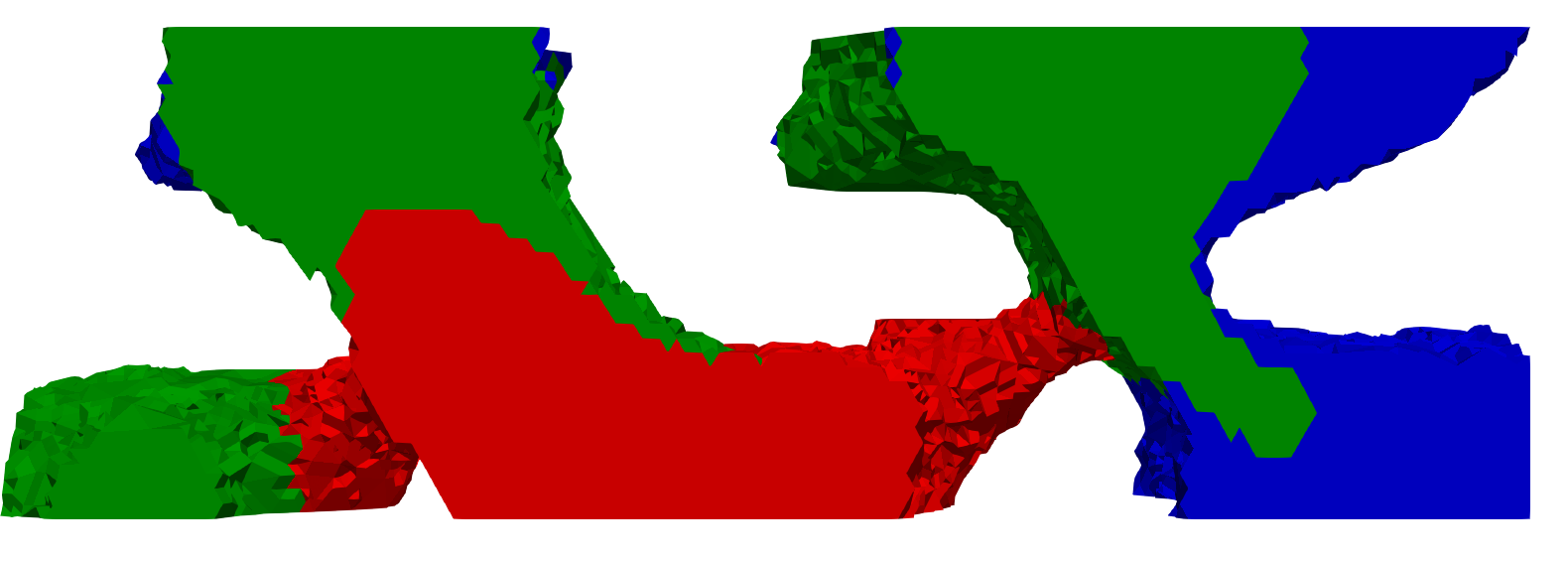}}
\subfigure[]{
\includegraphics[width=0.45\columnwidth]{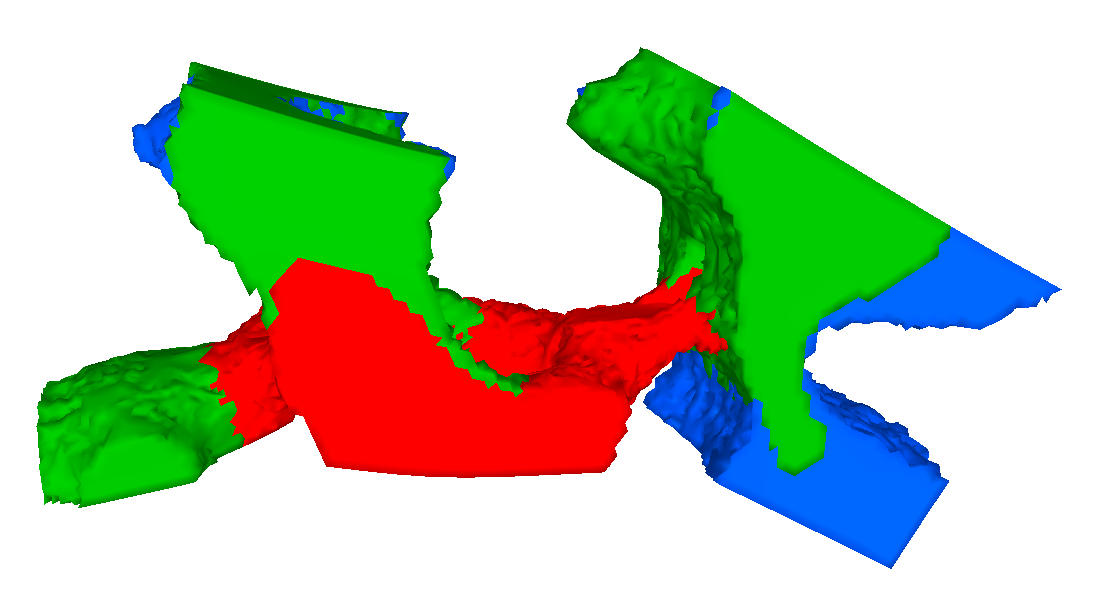}}
\subfigure[]{
\includegraphics[width=0.45\columnwidth]{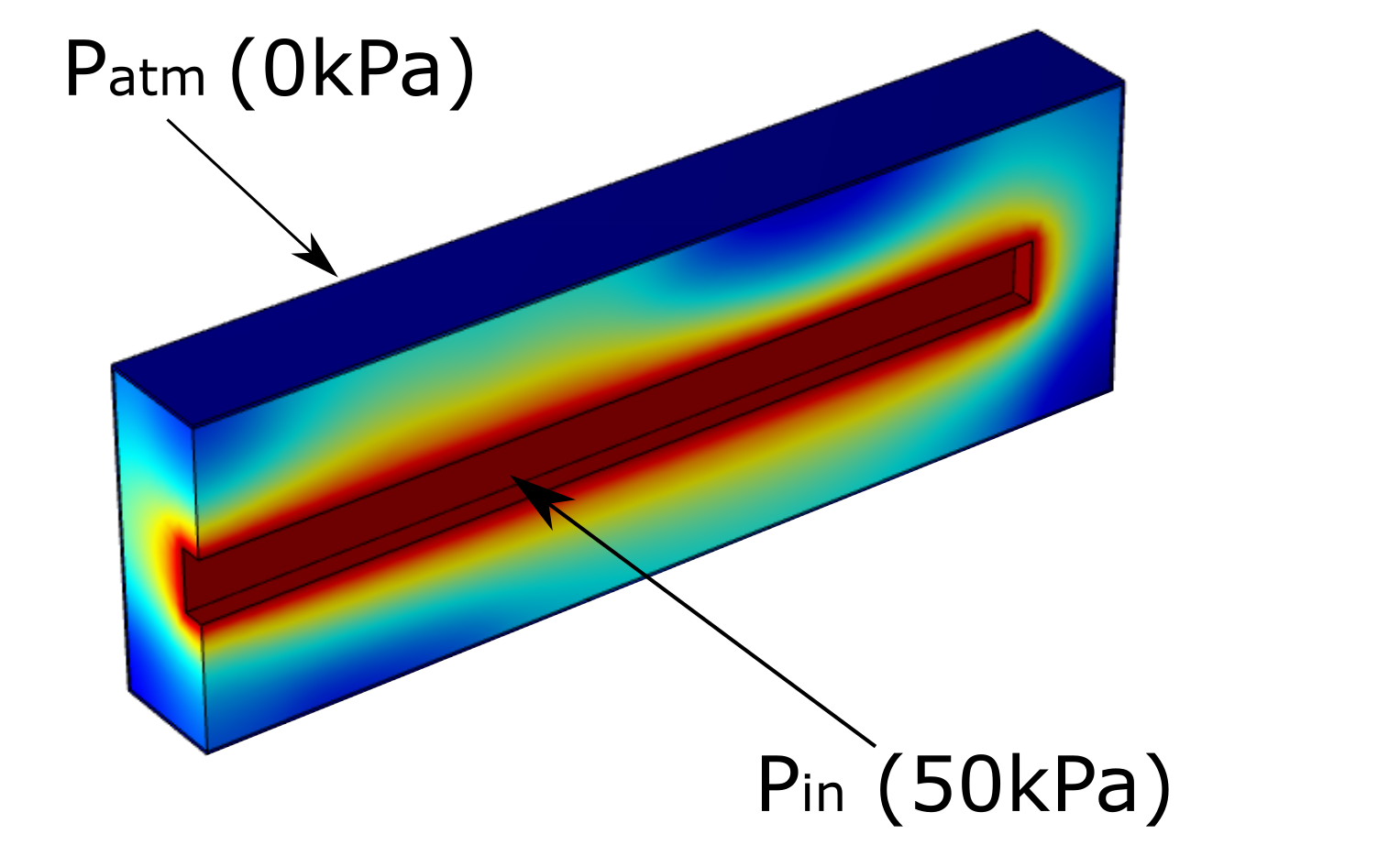}}
\subfigure[]{
\includegraphics[width=0.45\columnwidth]{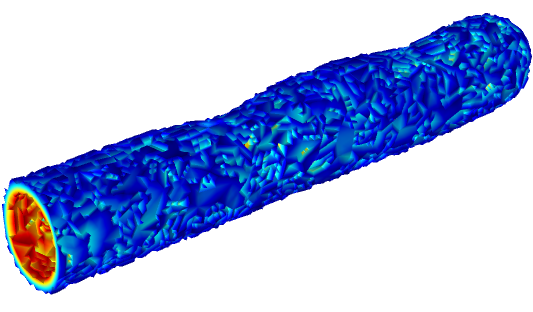}}
\subfigure[]{
\includegraphics[width=0.45\columnwidth]{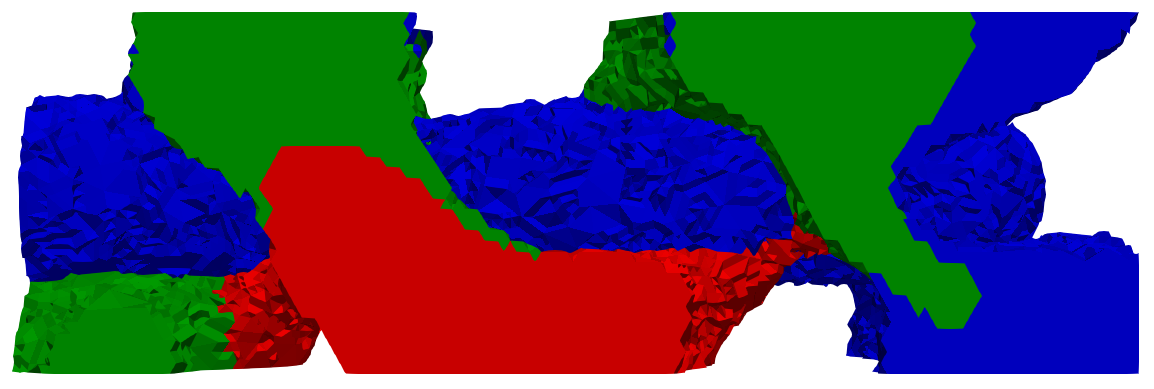}}
\caption{3-Material optimised soft-finger with stiffnesses: Red - 100MPa, Green - 10MPa, Blue - 1MPa,(a) Design domain  (b) Undeformed  (c) Deformed (5x deformation scale) (d) Optimised Pressure Distribution (Undeformed) (e) Implied pressure boundary (f) Complete design with sealed chamber }
\label{fig:Soft-Finger}
\end{figure}

An example of the design of the soft bending finger is shown in Figure \ref{fig:Soft-Finger}.
Without any closure method, the material is distributed roughly from stiffest to softest, with the stiffest material placed around the fixed side. Bending is increased by placing holes at the top and sides of the structure.
However, a closed structure is easily regenerated using the heuristic method.

\subsection{With Skin}
\begin{figure}[h!]
\centering
\subfigure[]{
\includegraphics[width=0.45\columnwidth]{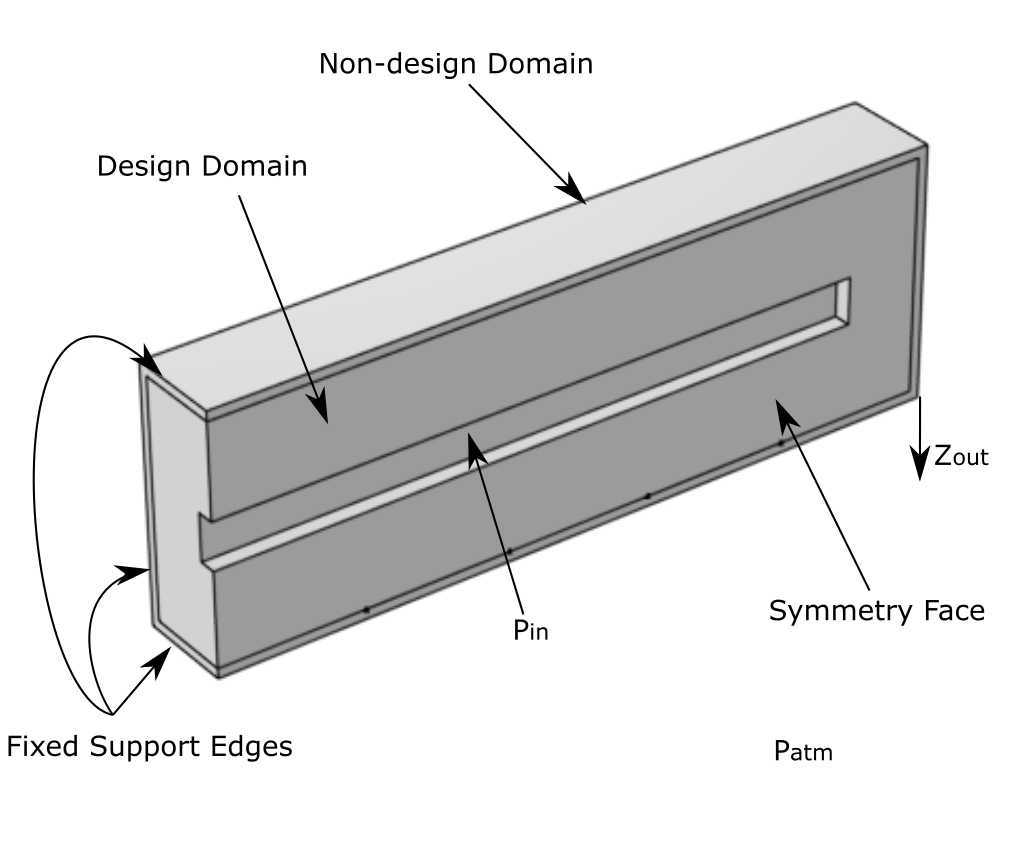}}
\subfigure[]{
\includegraphics[width=0.45\columnwidth]{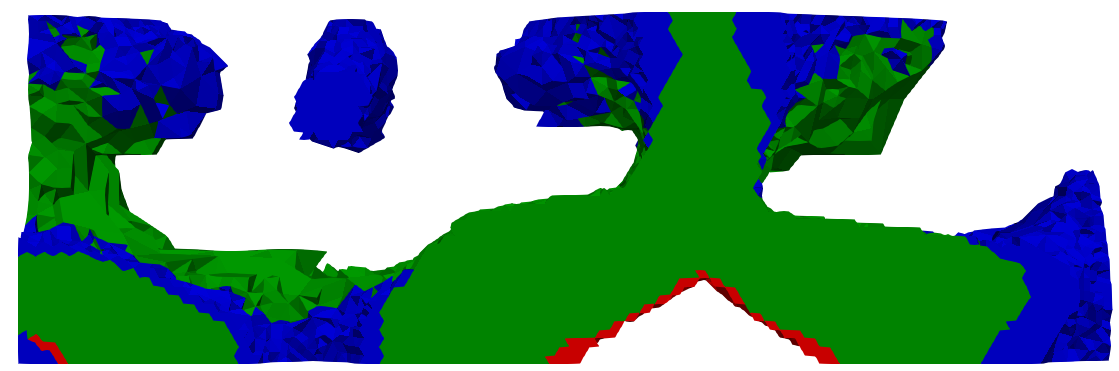}}
\subfigure[]{
\includegraphics[width=0.45\columnwidth]{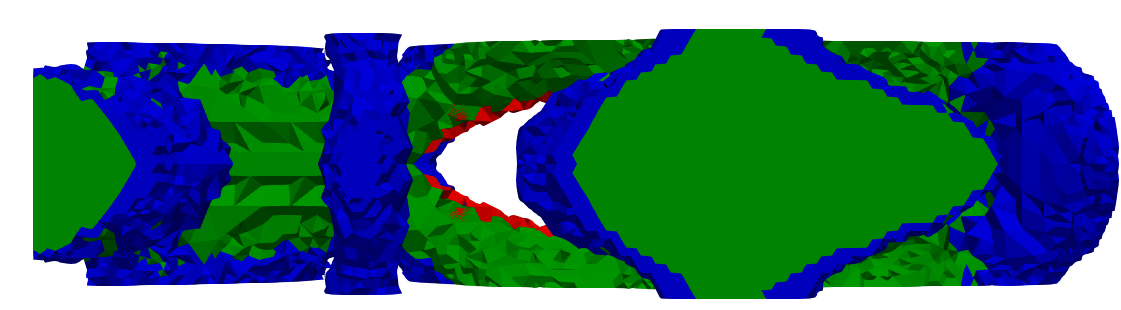}}
\subfigure[]{
\includegraphics[width=0.45\columnwidth]{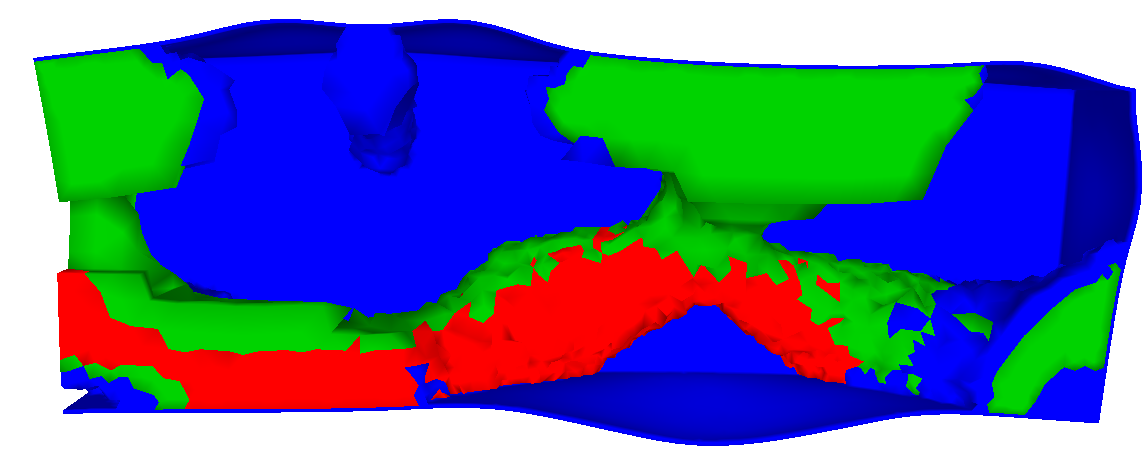}}
\caption{3-Material optimised soft-finger with casing - stiffnesses: Red - 100MPa, Green - 10MPa, Blue - 1MPa, (a) Design domain  (b) Undeformed Side-view (casing not shown) (c) Undeformed top-view (casing not shown) (d) Deformed half model, showing casing (5x deformation scale) }
\label{fig:Soft-Finger_skin}
\end{figure}

The surface can also be inserted as part of the optimisation problem by creating a non-design domain on the boundary of the optimisation region and assigning it to have stiffness $E_1$. This guarantees air cannot leak, but will produce suboptimal solutions as the external boundary must bend and expand to generate deformation far from the neutral bending axis (Figure \ref{fig:Soft-Finger_skin}). In contrast, Pneunets, a state of the art design have a sinusoidal profile which localises bending in narrow sections.

\subsection{Energy Penalty}
\begin{figure}[h!]
\centering
\subfigure[]{
\includegraphics[width=0.45\columnwidth]{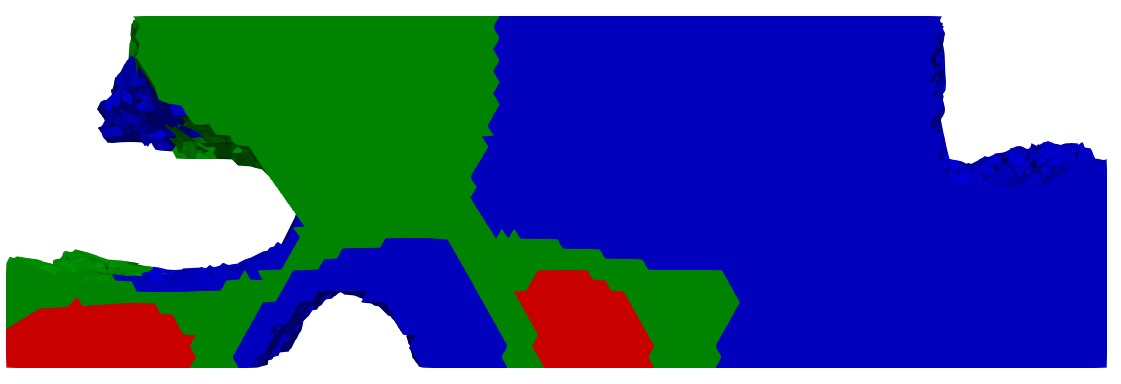}}
\subfigure[]{
\includegraphics[width=0.45\columnwidth]{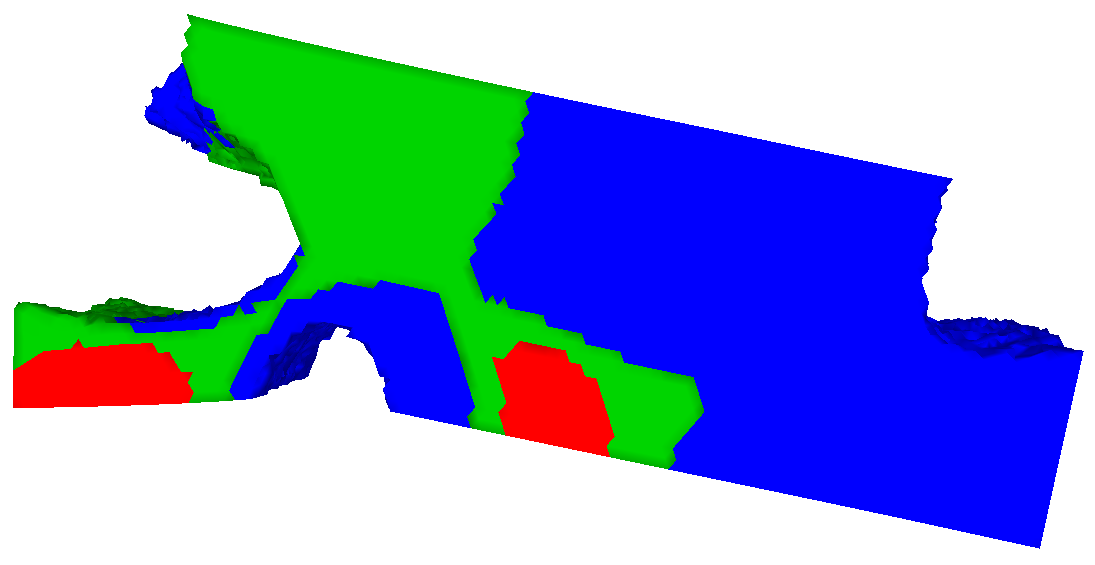}}
\caption{Energy Penalised soft-gripper(a) Undeformed (b) Deformed (5x deformation scale) }
\label{fig:Soft-Finger_energycost_v1=0.3}
\end{figure}

\begin{figure}[h!]
\centering
\subfigure[]{
\includegraphics[width=0.45\columnwidth]{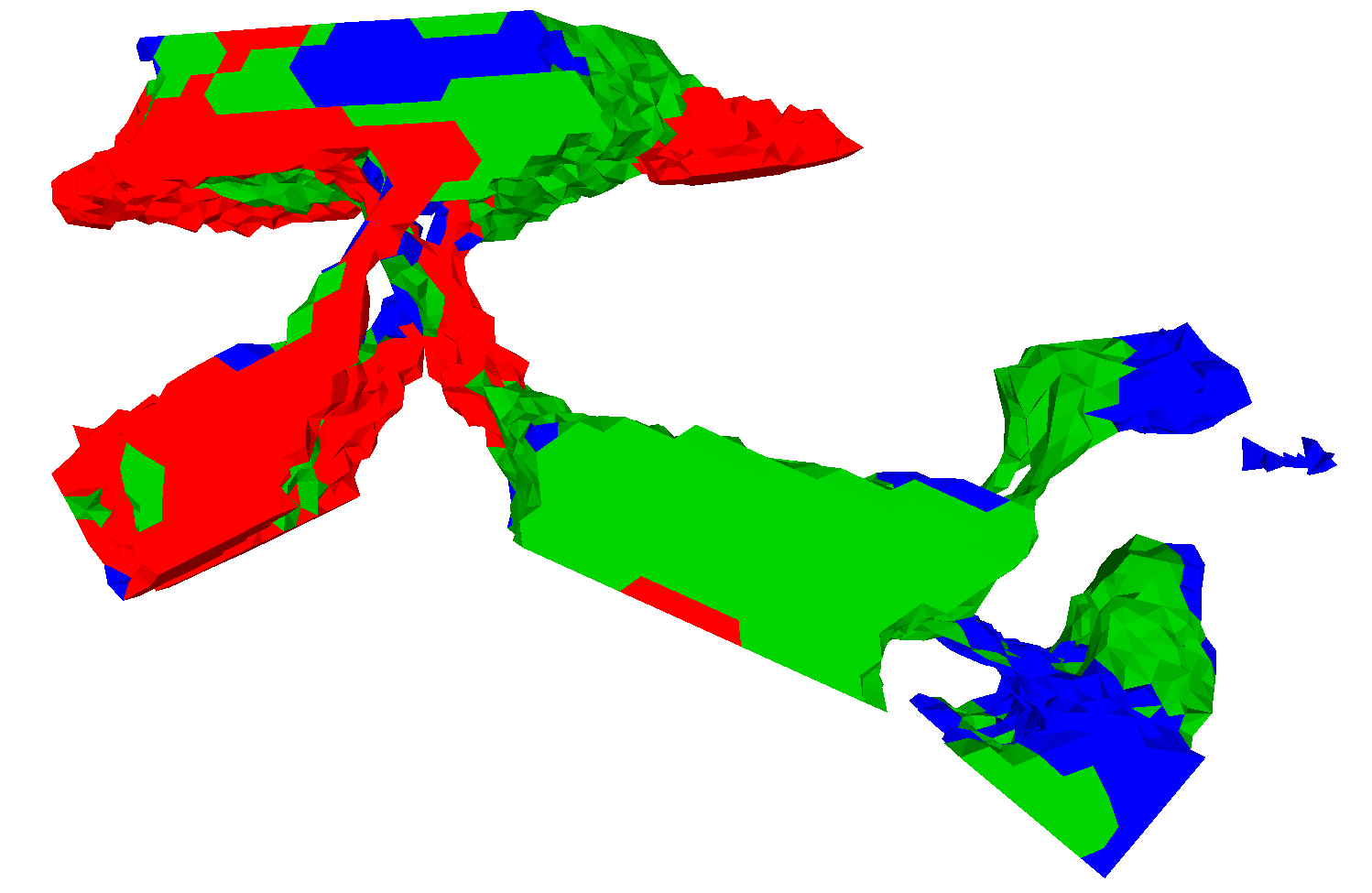}}
\subfigure[]{
\includegraphics[width=0.45\columnwidth]{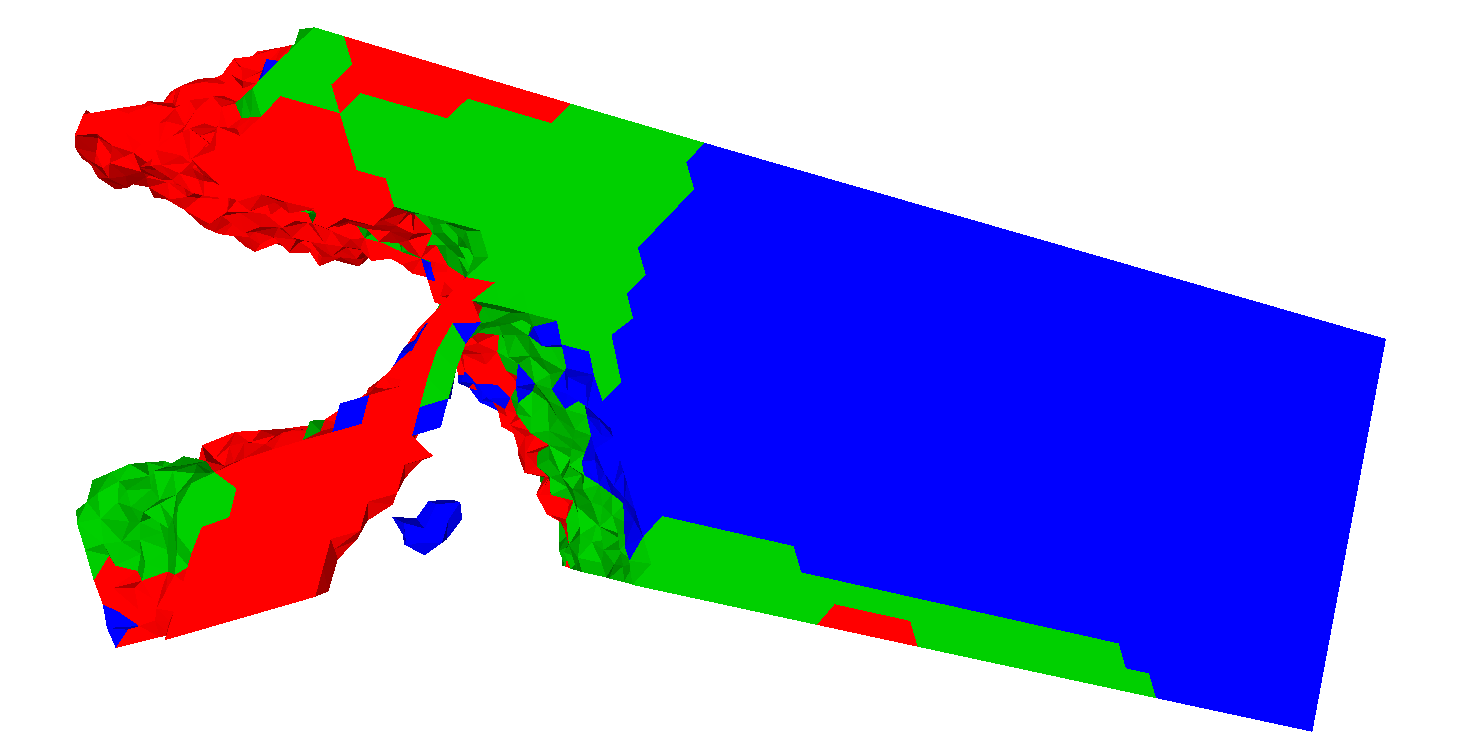}}
\caption{Energy Penalised soft-gripper with (a) $V_{f_1}=0.2$ (b) $V_{f_1}=0.4$}
\label{fig:Soft-Finger_energycost}
\end{figure}

Finally, the same design is presented using the energy penalty method. Here, the optimiser has reduced the overall amount of air leakage by using the low stiffness material $E_1$ to close sections of the chamber which contribute least to bending. As shown in Figure \ref{fig:Soft-Finger_energycost_v1=0.3}, the result is not a totally closed design, but one where the open areas have been greatly reduced. This uses the same design domain as the heuristic skin.
The efficacy of this penalty can be increased by increasing the volume limit of the most elastic material $E_1$. This is illustrated in Figure \ref{fig:Soft-Finger_energycost} in which the energy penalty is evaluated with $V_{f_1}=0.2$ and $V_{f_1}=0.4$. When using $V_{f_1}=0.2$, there is insufficient material to meaningfully close the design, but at $V_{f_1}=0.4$ an almost sealed chamber emerges with only a small opening around the fixed side.

\subsection{Numerical Comparison}
\begin{figure*}[h!]
\centering
\subfigure[]{
\includegraphics[width=0.95\columnwidth]{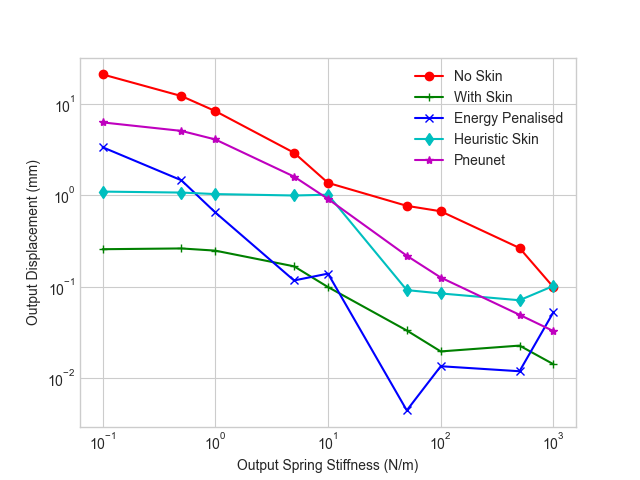}}
\subfigure[]{
\includegraphics[width=0.95\columnwidth]{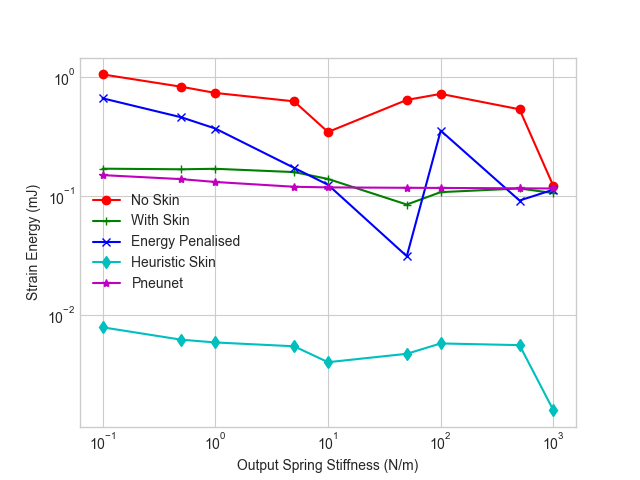}}
\subfigure[]{
\includegraphics[width=0.95\columnwidth]{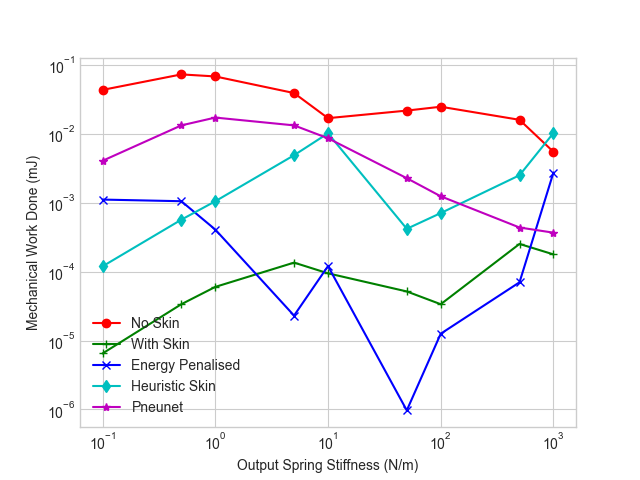}}
\subfigure[]{
\includegraphics[width=0.95\columnwidth]{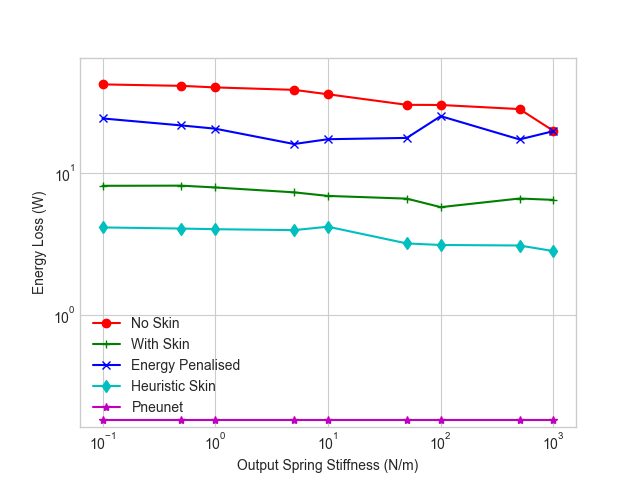}}
\caption{Energy Penalised soft-gripper with $V1=0.4$ (a) Displacement (b) Strain Energy (c) Mechanical work exerted on spring (d) Energy Loss }
\label{fig:Results_Plot}
\end{figure*}

We compare the two proposed closure methods by calculating their output displacement, strain energy, mechanical work done, and energy loss across 9 different output stiffnesses (springs placed at the output face) from $\SI{0.1}{\newton\per\meter}$ to $\SI{1000}{\newton\per\meter}$. The results are presented in Figure \ref{fig:Results_Plot}. A standard Pneunet design is included for comparison. It contains 7 inflatable chambers with a rectangular cross section, has total dimensions $\SI{17}{\milli\meter}$ x $\SI{15}{\milli\meter}$ x $\SI{72}{\milli\meter}$ and is made of a single material with $E=E_1=\SI{1}{\mega\pascal}$ and a constant wall thickness of $\SI{15}{\milli\meter}$.
Unsurprisingly, the unconstrained (no skin) optimisation produces the greatest bending, strain energy, work done and energy loss. 
Ignoring the energy loss, the design performs extremely well. In contrast, the closed design domain performs poorly. The heuristic gives the best performance of the methods discussed in this work, with a relatively large output displacement and low strain energy and energy loss. Of the methods discussed in this work, the heuristic gives the best performance, with a relatively large output displacement and low strain energy and energy loss. Whilst the energy penalty shows promise it is impeded by minimum length scales of topology optimisation, which prevent the formation of thin skins, and tends to become trapped in suboptimal local minima.
In contrast, the Pneunet design gives a relatively large displacement across the entire range of output stiffnesses and can exert a significant amount of work on the output spring, but to do so it must take up large amounts of internal strain. This inherent softness is beneficial when acting in free space or on very soft objects but detrimental when grasping stiffer ones as its output work declines at higher output stiffnesses.

\subsection{Experimental Validation}
\begin{figure*}[h!]
\centering
\includegraphics[width=2\columnwidth]{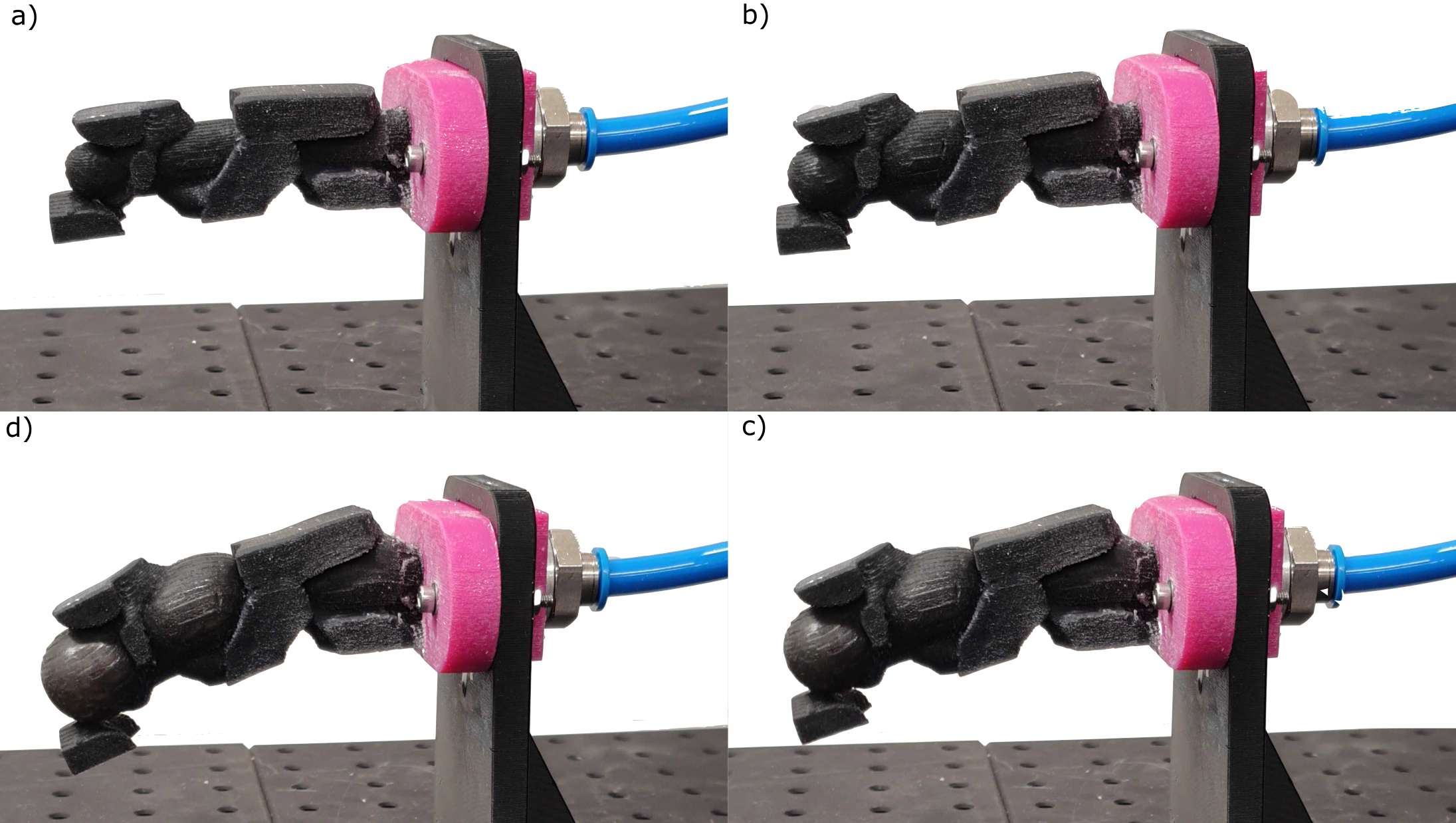}
\caption{Topology Optimised 3 material soft finger during inflation from (a) 0kPa to (d)100kPa}
\label{fig:PrintedSoftGripper}
\end{figure*}

To validate the concept, a heuristic skin soft finger was 3D printed using a Stratasys Connex3 Polyjet printer. It allows blending of multiple base materials to produce soft elastomers ranging from Shore-A 30 to 95 as well as rigid materials. The three optimised materials are approximated as Shore-A 30, Shore-A 60 and Shore-A 85. Figure \ref{fig:PrintedSoftGripper} shows the resulting printed finger in its undeformed state and during inflation. Although the printed material properties are only an approximation of the optimisation materials, the qualitative behaviour of the actuator matches the optimisation, validating the modelling and design approach. In the future, the mechanical properties of the 3D printed materials will be characterised to enhance simulation accuracy further.

\section{Conclusion}
\label{sec:conc}
Guaranteeing closure in the topology optoimisation of pneumatically actuated soft robots is a significant problem which has not been solved in existing research. We discussed two new methods for generated airtight or low-leakage soft robots. Of the two, the heuristic approach outperforms the rigorous optimisation method, but the latter approach is worthy of further investigation. In addition, this work presented a multi-material method for pneumatic topology optimisation and several new soft gripper designs. In numerical studies, the best optimised designs performed comparably to a Pneunet, a state of the art single material design. However, the topology optimisation method shows promise in generating bespoke designs for specific grasping challenges and can be generalised to any problem involving soft robotic motion. Nonlinearities including large-deformation, hyperelasticity and contact remain a challenge in topology optimisation. In future we aim to experimentally validate our designs in soft grasping and investigate the optimisation of non-linear mechanics. 

\bibliographystyle{hieeetr}
\bibliography{Referencesfinal}

\begin{thebibliography}{10}

\bibitem{Hao2021}
Y.~Hao and Y.~Visell, ``{Beyond Soft Hands: Efficient Grasping With
  Non-Anthropomorphic Soft Grippers},'' {\em Frontiers in Robotics and AI},
  vol.~8, no.~July, pp.~1--8, 2021.

\bibitem{Mosadegh2014}
B.~Mosadegh, P.~Polygerinos, C.~Keplinger, S.~Wennstedt, R.~F. Shepherd,
  U.~Gupta, J.~Shim, K.~Bertoldi, C.~J. Walsh, and G.~M. Whitesides,
  ``{Pneumatic Networks for Soft Robotics that Actuate Rapidly},'' {\em
  Advanced Functional Materials 24}, vol.~24, no.~15, pp.~2163--2170, 2014.

\bibitem{Laschi2012}
C.~Laschi, M.~Cianchetti, B.~Mazzolai, L.~Margheri, M.~Follador, and P.~Dario,
  ``{Soft robot arm inspired by the octopus},'' {\em Advanced Robotics},
  vol.~26, no.~7, pp.~709--727, 2012.

\bibitem{Brown2010}
E.~Brown, N.~Rodenberg, J.~Amend, A.~Mozeika, E.~Steltz, M.~R. Zakin,
  H.~Lipson, and H.~M. Jaeger, ``{Universal robotic gripper based on the
  jamming of granular material},'' {\em Proceedings of the National Academy of
  Sciences of the United States of America}, vol.~107, no.~44,
  pp.~18809--18814, 2010, 1009.4444.

\bibitem{Howard2021}
G.~D. Howard, J.~Brett, J.~O'Connor, J.~Letchford, and G.~W. Delaney,
  ``{One-Shot 3D-Printed Multimaterial Soft Robotic Jamming Grippers},'' {\em
  Soft Robotics}, vol.~00, no.~00, pp.~1--12, 2021.

\bibitem{Pinskier2022}
J.~Pinskier and D.~Howard, ``{From Bioinspiration to Computer Generation:
  Developments in Autonomous Soft Robot Design},'' {\em Advanced Intelligent
  Systems}, vol.~4, no.~1, p.~2100086, 2022.

\bibitem{Howard2022a}
D.~Howard, J.~O'Connor, J.~Letchford, J.~Brett, T.~Joseph, S.~Lin, D.~Furby,
  and G.~W. Delaney, ``{Getting a Grip: in Materio Evolution of Membrane
  Morphology for Soft Robotic Jamming Grippers},'' {\em 2022 IEEE 5th
  International Conference on Soft Robotics, RoboSoft 2022}, pp.~531--538,
  2022, 2111.01952.

\bibitem{Fitzgerald2021}
S.~G. Fitzgerald, G.~W. Delaney, D.~Howard, and F.~Maire, ``{Evolving soft
  robotic jamming grippers},'' {\em GECCO2021}, pp.~102--110, 2021.

\bibitem{Hiller2009}
J.~D. Hiller and H.~Lipson, ``{Multi material topological optimization of
  structures and mechanisms},'' {\em Proceedings of the 11th Annual Genetic and
  Evolutionary Computation Conference, GECCO-2009}, pp.~1521--1528, 2009.

\bibitem{Auerbach2011}
J.~Auerbach and J.~Bongard, ``{Evolving complete robots with CPPN-NEAT : The
  utility of recurrent connections},'' in {\em GECCO 2011}, no.~January, 2011.

\bibitem{Kriegman2020a}
S.~Kriegman, A.~M. Nasab, D.~Shah, H.~Steele, G.~Branin, M.~Levin, J.~Bongard,
  and R.~Kramer-Bottiglio, ``{Scalable sim-to-real transfer of soft robot
  designs},'' in {\em 3rd IEEE International Conference on Soft Robotics
  (RoboSoft) Yale University, USA Scalable}, (Yale University), pp.~359--366,
  2020.

\bibitem{sigmund2013topology}
O.~Sigmund and K.~Maute, ``Topology optimization approaches,'' {\em Structural
  and Multidisciplinary Optimization}, vol.~48, no.~6, pp.~1031--1055, 2013.

\bibitem{Chen2018a}
F.~Chen, W.~Xu, H.~Zhang, Y.~Wang, J.~Cao, M.~Y. Wang, H.~Ren, J.~Zhu, and
  Y.~Zhang, ``Topology optimized design, fabrication, and characterization of a
  soft cable-driven gripper,'' {\em IEEE Robotics and Automation Letters},
  vol.~3, no.~3, pp.~2463--2470, 2018.

\bibitem{kumar2020topology3Dpressure}
P.~Kumar and M.~Langelaar, ``On topology optimization of design-dependent
  pressure-loaded three-dimensional structures and compliant mechanisms,'' {\em
  International Journal for Numerical Methods in Engineering}, vol.~122, no.~9,
  pp.~2205--2220, 2021.

\bibitem{Liu2018}
C.~H. Liu, T.~L. Chen, C.~H. Chiu, M.~C. Hsu, Y.~Chen, T.~Y. Pai, W.~G. Peng,
  and Y.~P. Chiang, ``{Optimal design of a soft robotic gripper for grasping
  unknown objects},'' {\em Soft Robotics}, vol.~5, no.~4, pp.~452--465, 2018.

\bibitem{Zhang2019}
H.~Zhang, A.~S. Kumar, F.~Chen, J.~Y. Fuh, and M.~Y. Wang, ``{Topology
  optimized multimaterial soft fingers for applications on grippers,
  rehabilitation, and artificial hands},'' {\em IEEE/ASME Transactions on
  Mechatronics}, vol.~24, no.~1, pp.~120--131, 2019.

\bibitem{Liu2022}
C.~H. Liu, L.~J. Chen, J.~C. Chi, and J.~Y. Wu, ``{Topology Optimization Design
  and Experiment of a Soft Pneumatic Bending Actuator for Grasping
  Applications},'' {\em IEEE Robotics and Automation Letters}, pp.~1--8, 2022.

\bibitem{Chen2019a}
Y.~Chen, Z.~Xia, and Q.~Zhao, ``{Optimal Design of Soft Pneumatic Bending
  Actuators Subjected to Design-Dependent Pressure Loads},'' {\em IEEE/ASME
  Transactions on Mechatronics}, vol.~24, no.~6, pp.~2873--2884, 2019.

\bibitem{Caasenbrood2020}
B.~Caasenbrood, A.~Pogromsky, and H.~Nijmeijer, ``{A Computational Design
  Framework for Pressure-driven Soft Robots through Nonlinear Topology
  Optimization},'' {\em 2020 3rd IEEE International Conference on Soft
  Robotics, RoboSoft 2020}, no.~July, pp.~633--638, 2020.

\bibitem{Chen2020}
F.~Chen and M.~Y. Wang, ``{Design Optimization of Soft Robots: A Review of the
  State of the Art},'' {\em IEEE Robotics and Automation Magazine},
  no.~December, pp.~27--43, 2020.

\bibitem{Sigmund2007a}
O.~Sigmund and P.~M. Clausen, ``{Topology optimization using a mixed
  formulation: An alternative way to solve pressure load problems},'' {\em
  Computer Methods in Applied Mechanics and Engineering}, vol.~196, no.~13-16,
  pp.~1874--1889, 2007.

\bibitem{kumar2020topology}
P.~Kumar, J.~Frouws, and M.~Langelaar, ``Topology optimization of fluidic
  pressure-loaded structures and compliant mechanisms using the {Darcy}
  method,'' {\em Structural and Multidisciplinary Optimization}, vol.~61,
  pp.~1637--1655, 2020.

\bibitem{Desouza2020}
E.~M. de~Souza and E.~C.~N. Silva, ``{Topology optimization applied to the
  design of actuators driven by pressure loads},'' {\em Structural and
  Multidisciplinary Optimization}, vol.~61, no.~5, pp.~1763--1786, 2020.

\bibitem{sigmund1997design}
O.~Sigmund and S.~Torquato, ``Design of materials with extreme thermal
  expansion using a three-phase topology optimization method,'' {\em Journal of
  the Mechanics and Physics of Solids}, vol.~45, no.~6, pp.~1037--1067, 1997.

\bibitem{bruns2001topology}
T.~E. Bruns and D.~A. Tortorelli, ``Topology optimization of non-linear elastic
  structures and compliant mechanisms,'' {\em Computer methods in applied
  mechanics and engineering}, vol.~190, no.~26-27, pp.~3443--3459, 2001.

\end{thebibliography}

\end{document}